\documentclass[]{servicenow}
\usepackage[utf8]{inputenc}
\usepackage{amsmath}
\usepackage{amsfonts}
\usepackage{nicefrac}
\usepackage{enumitem}
\usepackage{multirow}
\usepackage{url}
\usepackage{graphicx}
\usepackage{xcolor}
\usepackage{booktabs}

\definecolor{bestcolor}{HTML}{E8F5E9}

\title{StarHarness: Evolving Harnesses with Stratified Search for Enterprise Environments}
\author[1]{Esakkivel Esakkiraja}
\author[1]{Denis Akhiyarov}
\author[1]{Vikas Yadav}
\author[1,2,3]{Sai Rajeswar}
\author[1]{Patrice Bechard}
\author[1]{Sridhar Nemala}
\author[1]{Sagar Davasam}

\affiliation[1]{ServiceNow}
\affiliation[2]{Mila}
\affiliation[3]{Universit\'e de Montr\'eal}

\abstract{

We present StarHarness, a framework for evolving environment-specific agent harnesses while keeping model weights fixed.
The evolved harness can include prompt and task framing, tool interfaces, skills, MCP-backed providers, subagent structure, and agent-loop configuration.
StarHarness constructs a compact evolution pool by stratifying tasks by baseline failure behavior, separates proposer-visible search tasks from proposer-hidden selection tasks, and reserves held-out tasks for evaluating generalization.
Across ITBench SRE, EnterpriseOps-Gym ITSM, and AutomationBench Finance, harness evolution improves full-benchmark performance by 20--35 percentage points over the default harness, after 4--12 accepted changes per environment.
These gains persist on tasks excluded from evolution and transfer without re-evolution across GPT and Qwen model families.
Trace analysis links them to interface repairs, environment conventions, and operational knowledge that compresses search, with fewer false-positive diagnoses and shorter trajectories in several settings.
StarHarness therefore offers a practical way to reduce persistent model--environment mismatch in tool-rich enterprise tasks.
}

\correspondence{\email{esakkivel.esakkiraja@servicenow.com}, \email{denis.akhiyarov@servicenow.com}}
\github{\url{https://github.com/ServiceNow/StarHarness}}

\begin{document}
\maketitle

\section{Introduction}

Modern LLM agents act through harnesses that define how they use tools and interpret state. Tool-augmented agents depend on the interaction between reasoning, actions, and external state~\citep{yao2023react, schick2023toolformer}. In a tool-rich enterprise task, harness design determines whether the agent can retrieve the right records, perform valid mutations, and satisfy exact final-state checks~\citep{rombaut2026taxonomy, meng2026harness_survey}. Interface design can change agent behavior and task success without changing model weights~\citep{yang2024swe_agent}.

Enterprise service-management and workflow agents must act through stateful backends, large tool surfaces, cross-step dependencies, and domain conventions that tool schemas often omit. Our benchmarks capture different instances of this broader problem: ITBench tests root-cause analysis over operational telemetry~\citep{jha2025itbench}; EnterpriseOps-Gym tests state-changing ITSM workflows against database assertions~\citep{malay2026eops}; and AutomationBench tests multi-application finance workflows with programmatic state checks~\citep{shepard2026automationbench}. Together, they expose the state dependencies, policy constraints, and final-state requirements that make enterprise tool use difficult~\citep{yao2024taubench, lu2024toolsandbox}. They raise three questions: Can search evolve a harness from a compact task subset without overfitting to that subset? Do the resulting changes transfer to unseen tasks and other models? Which interaction failures does harness evolution repair?

StarHarness addresses these questions by evolving environment-specific harnesses while keeping model weights fixed. The method stratifies tasks by baseline failure behavior, separates proposer-visible search tasks from proposer-hidden selection tasks, and reserves held-out tasks for evaluation.

Our contributions:
\begin{enumerate}[leftmargin=*,itemsep=2pt]
\item \textbf{Efficient stratified harness evolution.}
We introduce a harness-evolution protocol that searches over compact, failure-mode-stratified task subsets while separating proposer-visible search tasks, proposer-hidden selection tasks, and held-out evaluation.
This provides a direct measure of generalization.

\item \textbf{Task and model transfer of specialized harnesses.}
Across three stateful enterprise benchmarks, harnesses evolved with one model improve tasks excluded from evolution and transfer without re-evolution across GPT and Qwen models.

\item \textbf{Analysis of learned specialization.}
We identify three recurring forms of environment specialization: interface repair, environment conventions, and operational knowledge that compresses search. We connect them to changes in agent precision, convergence, and efficiency.
\end{enumerate}

\section{Related Work}

\subsection{Prompt and Harness Optimization}

Prompt optimization methods search instructions and demonstrations with generated candidates, textual feedback, or evolutionary selection~\citep{opsahlong2024mipro, agrawal2025gepa}. Agent-architecture methods extend the search to executable modules and workflow structure~\citep{khattab2024dspy, zhang2025aflow}.

Recent harness-level systems edit executable scaffolding or co-evolve harnesses with model policies and weights~\citep{lee2026metaharness, hebbar2026sia}.

StarHarness studies a narrower deployment problem: adapting a frozen model's harness to a stateful enterprise environment. Its search space includes prompts, tools, skills, MCP providers, subagents, and execution policy. Evaluation protocols differ across this literature: Meta-Harness uses held-out sets for text classification and mathematical reasoning, but searches and reports final TerminalBench-2 performance on the same 89-task benchmark, which its authors frame as a discovery setting~\citep{lee2026metaharness}. We use task-level separation, proposer-hidden selection, and held-out evaluation to distinguish search performance from generalization, following recent calls for stricter harness-evolution evaluation~\citep{wang2026rethinking_harness}. The resulting edits remain external to model weights and can be tested and reverted as ordinary code changes.

\subsection{Agent Benchmarks and Enterprise Environments}

Existing benchmarks cover related agent settings with different interaction interfaces. WorkArena~\citep{drouin2024workarena} studies browser-based knowledge work on the ServiceNow platform, while Terminal-Bench 2.0~\citep{merrill2026terminalbench} evaluates difficult command-line tasks in isolated environments. We focus on three complementary enterprise settings: ITBench~\citep{jha2025itbench} contains 40 Kubernetes root-cause analysis scenarios in which the agent inspects alerts, events, traces, metrics, and topology before producing a structured diagnosis; EnterpriseOps-Gym~\citep{malay2026eops} contains 103 ITSM workflows spanning incident, problem, change, knowledge, and user-management tasks, with SQL verifiers checking the resulting ServiceNow state; and AutomationBench~\citep{shepard2026automationbench} contains 100 finance workflows across 47 simulated SaaS applications, scored by programmatic assertions over environment state. These benchmarks require agents to operate domain tools, maintain persistent state, and satisfy workflow-specific objectives. We use them to measure task-level generalization and transfer of a frozen specialized harness across models.

\section{Method}

\subsection{Overview}

We define \textbf{harness evolution} as outer-loop optimization of the executable scaffold surrounding a fixed language model. In our implementation, the \textbf{StarHarness optimizer} is a coding harness built on Oh My Pi~\citep{ohmypi2026}, a variant of the Pi agent harness~\citep{zechner2026pi}. It hosts the proposer and executes the edit, validation, and evaluation loop: it exposes the permitted repository and benchmark traces, applies a candidate patch, runs checks, and launches benchmark evaluations. The optimizer modifies the separate \textbf{Stirrup harness}~\citep{artificialanalysis2026stirrup}, which serves as the agent harness under evaluation. Its editable surface includes prompt and task framing, tool definitions and schemas, argument preprocessing, skills, MCP providers, subagent structure, context management, verification, and finish logic. The model weights and benchmark remain fixed during evolution.

Let $\mathcal{D}$ denote a benchmark, $h$ a harness, $M$ the agent model, and $J(h;\mathcal{D})$ the cost function, defined here as the mean task score obtained by running $M$ with $h$ on $\mathcal{D}$; higher values are better. The target is
\begin{equation}
h^* = \underset{h\in\mathcal{H}}{\arg\max}\; J(h;\mathcal{D}_{\mathrm{holdout}}),
\label{eq:objective}
\end{equation}
where $\mathcal{H}$ is the permitted harness space. StarHarness cannot access held-out outcomes during search, so it approximates this target with proposer-visible search tasks and proposer-hidden selection tasks where applicable, reserving $\mathcal{D}_{\mathrm{holdout}}$ for final evaluation.

\begin{figure*}[t]
\centering
\includegraphics[width=0.90\textwidth]{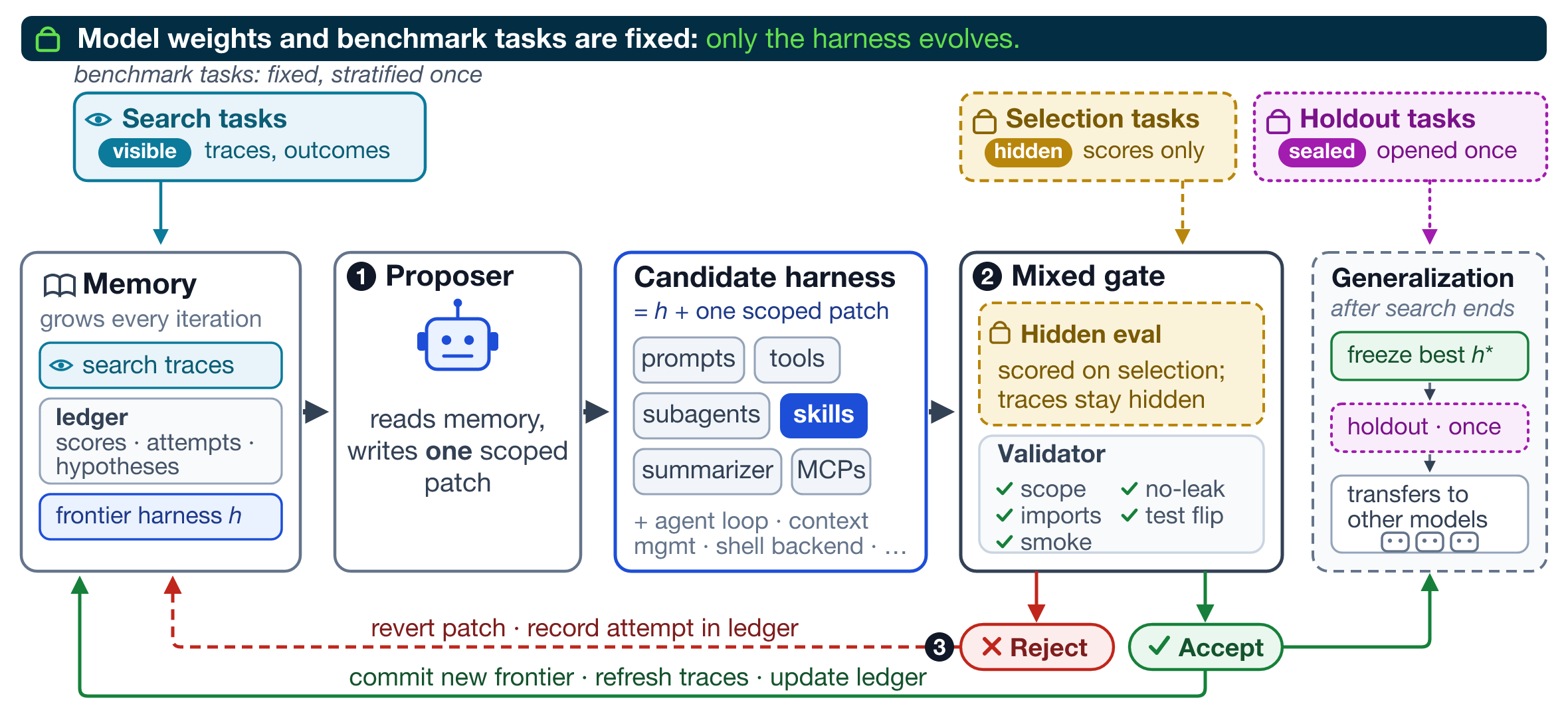}

\vspace{4pt}
\fbox{\parbox{0.68\textwidth}{\scriptsize
\textbf{Algorithm 1: StarHarness evolution} \\[2pt]
\textbf{Input:} benchmark $\mathcal{D}$; fixed model $M$; seed harness $h_0$; proposer $P$; budget $B$ \\
\textbf{Output:} evolved harness $h^*$ \\
\makebox[2.2em][l]{\texttt{1:}}Run $h_0$ on $\mathcal{D}$; record scores and failure strata \\
\makebox[2.2em][l]{\texttt{2:}}Construct $D_{\mathrm{search}},D_{\mathrm{select}},D_{\mathrm{holdout}}$ \\
\makebox[2.2em][l]{\texttt{3:}}$h \leftarrow h_0$; $s \leftarrow J(h;D_{\mathrm{select}})$; initialize ledger $L$ \\
\makebox[2.2em][l]{\texttt{4:}}\textbf{for} $t=1,\ldots,B$ \textbf{do} \\
\makebox[2.2em][l]{\texttt{5:}}\hspace*{1.2em}$L\leftarrow\text{load ledger}$; $T\leftarrow\text{gather search traces}$ \\
\makebox[2.2em][l]{\texttt{6:}}\hspace*{1.2em}$\Delta \leftarrow P(h,T,L)$; capture $\Delta$ as a scoped patch \\
\makebox[2.2em][l]{\texttt{7:}}\hspace*{1.2em}\textbf{if} scope, leakage, import, or smoke check fails: revert and record rejection \\
\makebox[2.2em][l]{\texttt{8:}}\hspace*{1.2em}\textbf{else} run proposer-selected test flip \\
\makebox[2.2em][l]{\texttt{9:}}\hspace*{2.4em}\textbf{if} test flip fails: revert and record rejection \\
\makebox[2.2em][l]{\texttt{10:}}\hspace*{2.4em}\textbf{else} evaluate $h\oplus\Delta$ on hidden $D_{\mathrm{select}}$ \\
\makebox[2.2em][l]{\texttt{11:}}\hspace*{3.6em}\textbf{if} selection score improves, or ties with an available verifier metric that improves: commit $h\oplus\Delta$; update $L$; refresh $T$ \\
\makebox[2.2em][l]{\texttt{12:}}\hspace*{3.6em}\textbf{else}: revert $\Delta$; update $L$ \\
\makebox[2.2em][l]{\texttt{13:}}\textbf{return} $h$; evaluate once on $D_{\mathrm{holdout}}$ and $\mathcal{D}$
}}
\caption{StarHarness evolution. \textit{Top:} the corresponding workflow. \textit{Bottom:} the executable procedure. The proposer reads the current frontier, persistent memory ledger, and search-set traces, but never selection or holdout traces. Candidates pass scoped validation and a proposer-selected test flip before hidden selection evaluation; accepted edits advance the committed frontier and refresh the ledger and search traces. For tree search, the ledger stores candidate nodes and parent links rather than only one greedy frontier.}
\label{fig:pipeline}
\end{figure*}

The coding harness runs three stages: proposal, validation, and evaluation. The proposer reads the current harness and search-set traces and returns a candidate patch. The validator checks scope, imports, and a single-task smoke test. The evaluator runs valid candidates on the hidden selection set when one exists; the evolution loop applies the fixed acceptance rule. Following the autoresearch pattern~\citep{karpathy2026autoresearch}, the evolution loop measures a baseline, proposes one bounded intervention, evaluates it, retains only an improvement, and repeats from the resulting frontier. A persistent memory ledger links patch, session, and evaluation artifacts and carries frontier scores, per-task outcomes, accepted hypotheses, and discarded attempts into later iterations. For tree search, the ledger is a solution journal containing candidate nodes, parent links, scores, failure states, and promotion decisions. Figure~\ref{fig:pipeline} gives the procedure in algorithmic form.

The evolution loop first runs a proposer-selected single-task \emph{test flip}. If the candidate does not flip that task, the loop records a rejection and skips the expensive evaluation. A candidate that passes the gate is evaluated on the selection set; promotion remains deterministic because the candidate must improve the selection mean, with verifier rate used as a tie-breaker only when that additional metric is available. The loop commits accepted candidates as the new frontier and refreshes the search traces. It reverts rejected, invalid, and crashed candidates to the prior frontier.

\subsection{Task Partition and Stratified Sampling}
\label{sec:stratified}

We construct the task partition before evolution. From the full benchmark of $N$ tasks, we first retain $N'$ tasks with reproducible evaluation. We then sample an evolution pool of $K$ tasks, typically $K \approx N/2$, using three baseline descriptors:
\begin{itemize}[leftmargin=*,itemsep=1pt]
\item \textbf{Baseline failure mode} (e.g., wrong\_tool, context\_loss, missing\_evidence, premature\_conclusion)
\item \textbf{Baseline task score}
\item \textbf{Verifier pass rate}
\end{itemize}
We compute these descriptors from a baseline run on all $N'$ tasks before creating the partition. When the protocol uses a selection set, we split the evolution pool into proposer-visible search tasks and proposer-hidden selection tasks while matching their baseline score, failure-mode, and verifier-pass distributions. The proposer receives search-task traces and outcomes, but no selection-task contents, traces, verifier feedback, or per-task outcomes. The remaining $N'-K$ tasks form the holdout set and never affect proposal or acceptance.

\subsection{Search Strategies: Exploration and Exploitation}
\label{sec:search}

We use the same proposer, validator, evaluator, and acceptance score in two search procedures. In \textbf{hill climbing}, the state is a single frontier harness. At iteration $t$, the proposer emits one patch $\Delta_t$ from traces of the current frontier; the system keeps $h_t\oplus\Delta_t$ if and only if it strictly improves the selection score, or ties it while improving an available verifier metric, and otherwise restores $h_t$.

In \textbf{tree search}, the state is a set of candidate nodes. Each node stores a parent pointer, cumulative patch, search traces, validation status, and selection score. The proposer can \emph{explore} a failure pattern, \emph{draft} a patch, \emph{debug} a failed candidate, \emph{merge} two compatible nodes, or \emph{improve} an existing node. Valid nodes are scored on the same hidden selection set; the best surviving node becomes the frontier for subsequent refinement. This preserves alternative hypotheses instead of committing after the first accepted edit.

We use these modes as an exploration--exploitation control experiment on EnterpriseOps-Gym alone. Tree search explores alternative hypotheses; hill climbing then exploits the best tree frontier through bounded local edits. The stages are sequential, so this design describes how the modes complement each other rather than providing a causal head-to-head comparison.

\subsection{Guardrails and Candidate Isolation}

StarHarness performs benchmark-assisted environment adaptation: the proposer may inspect search-task trajectories and their evaluation outcomes to diagnose recurring failures, but guardrails prevent it from encoding task-specific solutions.

Each candidate is a git diff against the current frontier, scoped to the benchmark's editable directories and the shared agent framework.
Forbidden changes include:
\begin{itemize}[leftmargin=*,itemsep=1pt]
\item Branching on task IDs or hard-coded answers
\item Verifier or assertion content in agent prompts
\item Ground-truth tables or hidden-state access
\item Benchmark-specific answer mappings
\end{itemize}

These constraints target reusable environment behavior rather than individual task solutions.
The validator checks scope, imports, and a single-task smoke test before benchmarking.
The evolution loop reverts candidates that fail any check without running the selection evaluation.

\section{Experiments}

\subsection{Experimental Setup}

\textbf{Benchmarks.}
We evaluate on three enterprise benchmarks:

\begin{itemize}[leftmargin=*,itemsep=2pt]
\item \textbf{ITBench SRE}~\citep{jha2025itbench}: the latest public release of the ITBench-AA SRE set, consisting of 40 Kubernetes root-cause analysis scenarios from the OpenTelemetry demo application~\footnote{\url{https://huggingface.co/datasets/ArtificialAnalysis/ITBench-AA}}. The agent inspects an offline incident snapshot containing alerts, events, traces, metrics, and topology, then writes a structured diagnosis identifying the contributing entities. Our Stirrup baseline follows the documented setup where available, including the sandboxed code-execution environment, shell-based snapshot inspection, and structured JSON output.
\item \textbf{EnterpriseOps-Gym ITSM}~\citep{malay2026eops}: 103 ITSM oracle tasks (incident, problem, change, knowledge, and user management), a subset of the broader 1{,}150-task benchmark, against a ServiceNow MCP backend. Graded by SQL verifiers checking final database state.
\item \textbf{AutomationBench Finance}~\citep{shepard2026automationbench}: 100 finance workflow tasks (AP/AR, expenses, reporting, bookkeeping) across 47 simulated SaaS applications. Graded by programmatic assertions on environment state. We report the share of the domain's objectives the model achieved; a guardrail violation assigns the task a score of zero. Our Finance-100 subset and Stirrup agent harness differ from the benchmark paper's default setup, so scores are not directly comparable to those reported by~\citet{shepard2026automationbench}.
\end{itemize}

Across all three benchmarks, we score following the Artificial Analysis evaluation description~\footnote{\url{https://artificialanalysis.ai/evaluations/}}. For AutomationBench, this score is computed from the share of the domain's objectives the model achieved, with a guardrail violation assigning the task a score of zero.

\textbf{Models.}
We evolve harnesses using GPT-5.4 (medium reasoning) as the agent under test and GPT-5.4 as the proposer running inside the Pi-based coding harness~\citep{openai2026gpt54}. We then evaluate the same frozen evolved harness, without re-evolution, on additional GPT and Qwen models, including Qwen3.6~\citep{qwen2026qwen36}, for each benchmark.

\textbf{Baselines.}
For each benchmark, the baseline is the unmodified Stirrup agent framework with default prompts and tool configurations.
We additionally compare against the standalone Pi and Codex harnesses and GEPA prompt optimization applied on top of the Pi harness.

\subsection{Main Results}

\begin{figure*}[t]
\centering
\includegraphics[width=0.47\textwidth]{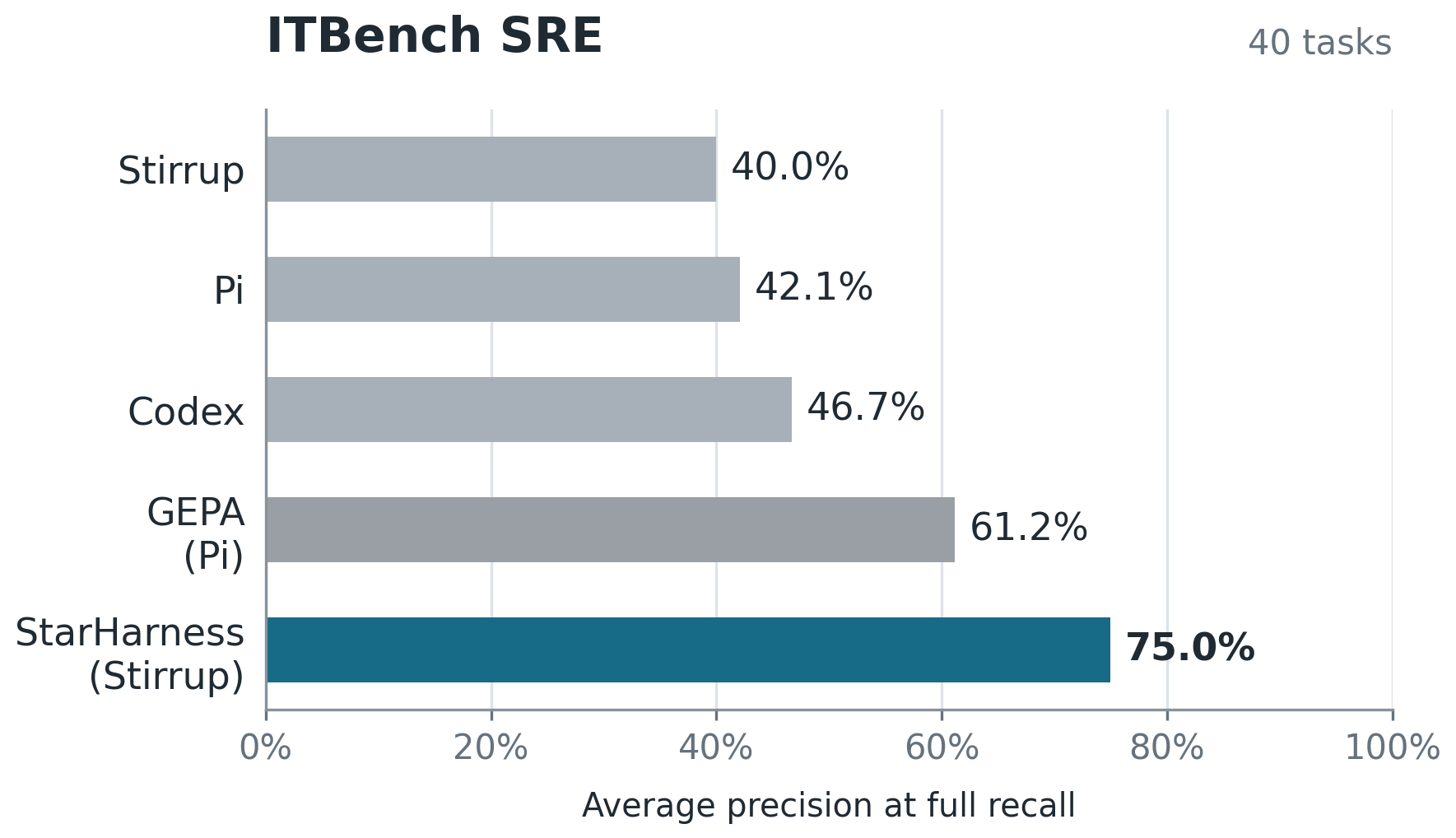}\hfill
\includegraphics[width=0.47\textwidth]{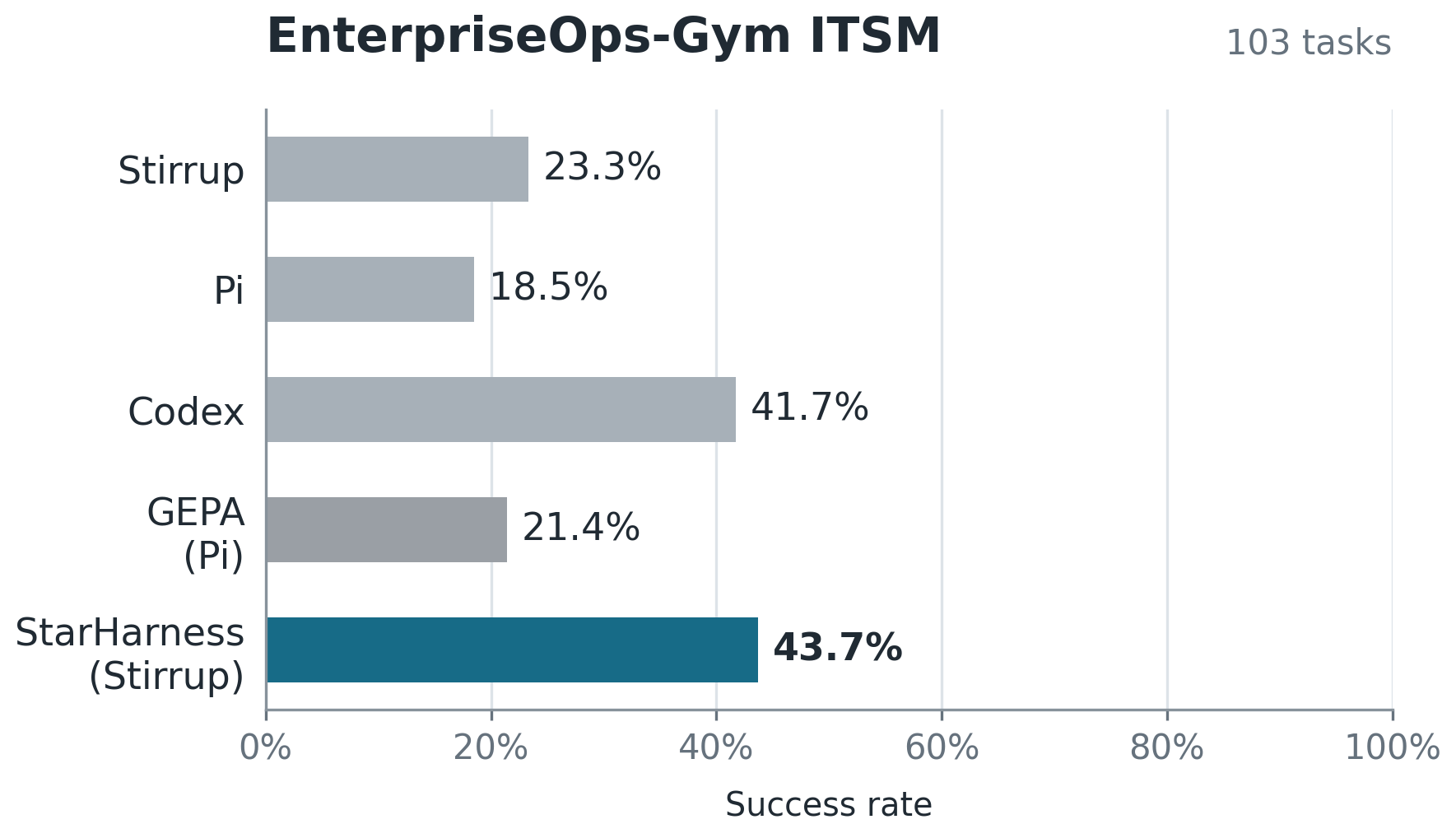}

\vspace{0.5em}
\includegraphics[width=0.47\textwidth]{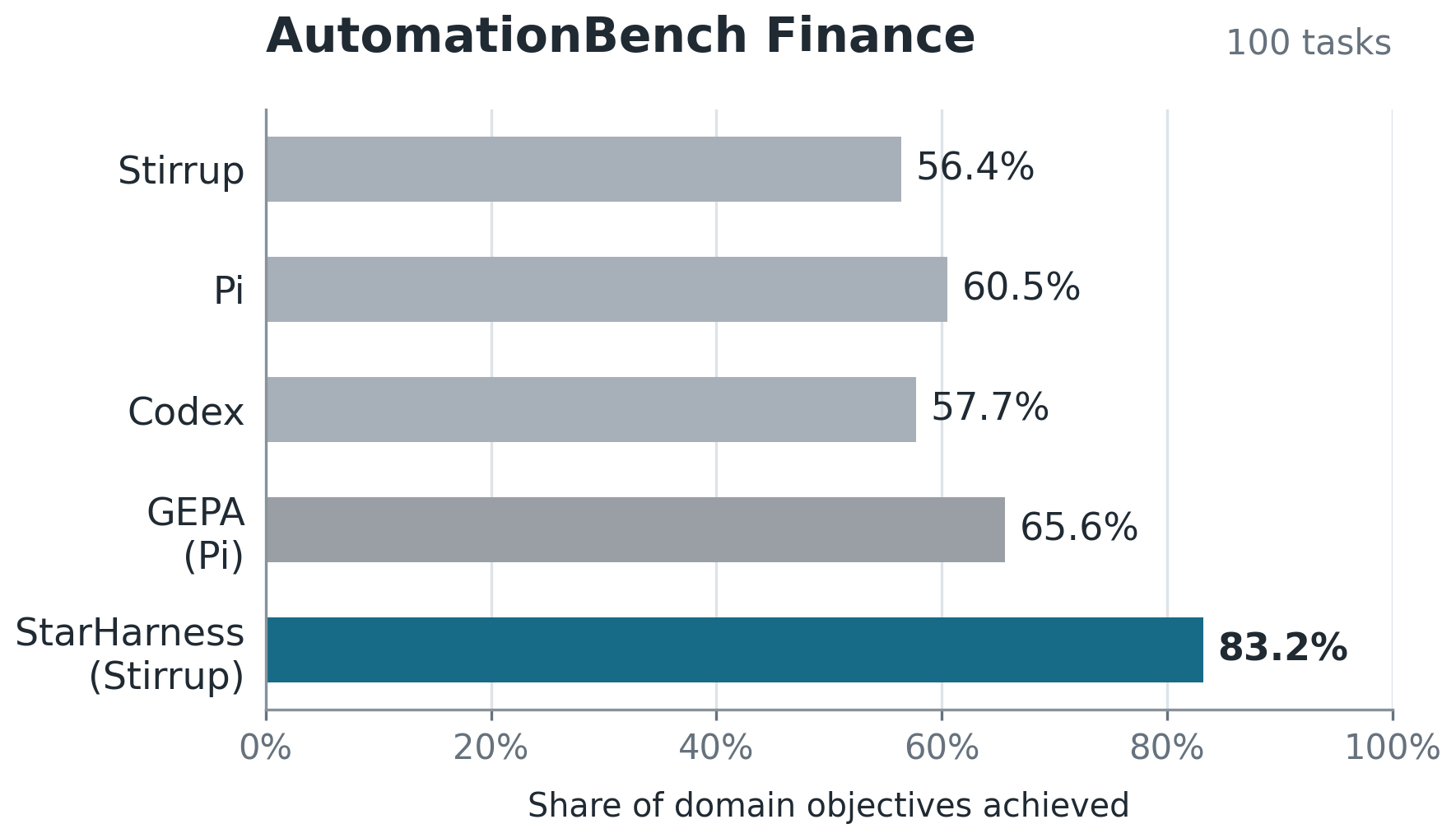}
\caption{Full-benchmark harness comparisons for ITBench SRE, EnterpriseOps-Gym ITSM, and AutomationBench Finance. GEPA (Pi) denotes prompt optimization applied on top of the Pi harness; StarHarness (Stirrup) denotes the evolved Stirrup harness. The AutomationBench score is the share of the domain's objectives the model achieved, with a guardrail violation assigning the task a score of zero.}
\label{fig:harness-comparison}
\end{figure*}

Figure~\ref{fig:harness-comparison} summarizes the full-benchmark harness comparison. Scores use each benchmark's metric; higher is better. StarHarness (Stirrup) is the evolved Stirrup harness, while GEPA (Pi) denotes prompt optimization applied on top of the Pi harness.

StarHarness (Stirrup) is the strongest configuration on all three benchmarks. Relative to GEPA (Pi)~\citep{agrawal2025gepa}, the gains are +13.8, +22.3, and +17.6 percentage points on ITBench, EnterpriseOps-Gym, and AutomationBench, respectively. The comparison is descriptive: the systems differ in prompts, tools, execution policies, and harness architecture, so it does not isolate a single causal component. The result shows that environment-specific harness design can add substantial performance beyond prompt optimization alone.

The performance gains also coincide with lower estimated GPT-5.4 inference cost per task: StarHarness reduces cost by 17\% on ITBench, 53\% on EnterpriseOps-Gym, and 29\% on AutomationBench at published rates~\citep{openai2026gpt54}.

\subsection{Frozen Cross-Model Transfer}

Table~\ref{tab:model-transfer} consolidates frozen-harness transfer results across all three benchmarks. Each evolved-harness score uses the same StarHarness artifact evolved with GPT-5.4; no benchmark-specific re-evolution was performed for the transferred models. The harness improves every transferred model in the table, including models from both GPT and Qwen families.

\begin{table*}[!htbp]
\caption{Frozen StarHarness transfer across models. Values are full-benchmark scores; parentheses give reasoning level when applicable.}
\label{tab:model-transfer}
\centering
\small
\begin{tabular}{@{}llrrr@{}}
\toprule
Benchmark & Model & Baseline & StarHarness & $\Delta$ \\
\midrule
ITBench & Qwen3.5-27B & 25.6\% & 70.0\% & +44.4 pp \\
ITBench & GPT-5.4-mini (medium) & 33.1\% & 79.4\% & +46.3 pp \\
ITBench & GPT-5.4 (medium) & 40.0\% & 75.0\% & +35.0 pp \\
ITBench & GPT-5.5 (medium) & 50.8\% & 78.7\% & +27.9 pp \\
\midrule
EnterpriseOps-Gym & Qwen3.6-27B & 18.2\% & 38.8\% & +20.6 pp \\
EnterpriseOps-Gym & GPT-5.4-mini (medium) & 13.6\% & 31.1\% & +17.5 pp \\
EnterpriseOps-Gym & GPT-5.4 (medium) & 23.3\% & 43.7\% & +20.4 pp \\
EnterpriseOps-Gym & GPT-5.5 (high) & 37.8\% & 48.5\% & +10.7 pp \\
\midrule
AutomationBench & Qwen3.6-27B & 48.2\% & 75.5\% & +27.3 pp \\
AutomationBench & GPT-5.4-mini (medium) & 29.6\% & 70.0\% & +40.4 pp \\
AutomationBench & GPT-5.4 (medium) & 57.1\% & 83.2\% & +26.1 pp \\
AutomationBench & GPT-5.5 (medium) & 59.6\% & 84.9\% & +25.3 pp \\
\bottomrule
\end{tabular}
\end{table*}
\FloatBarrier

\begin{figure}[t]
\centering
\includegraphics[width=0.96\linewidth]{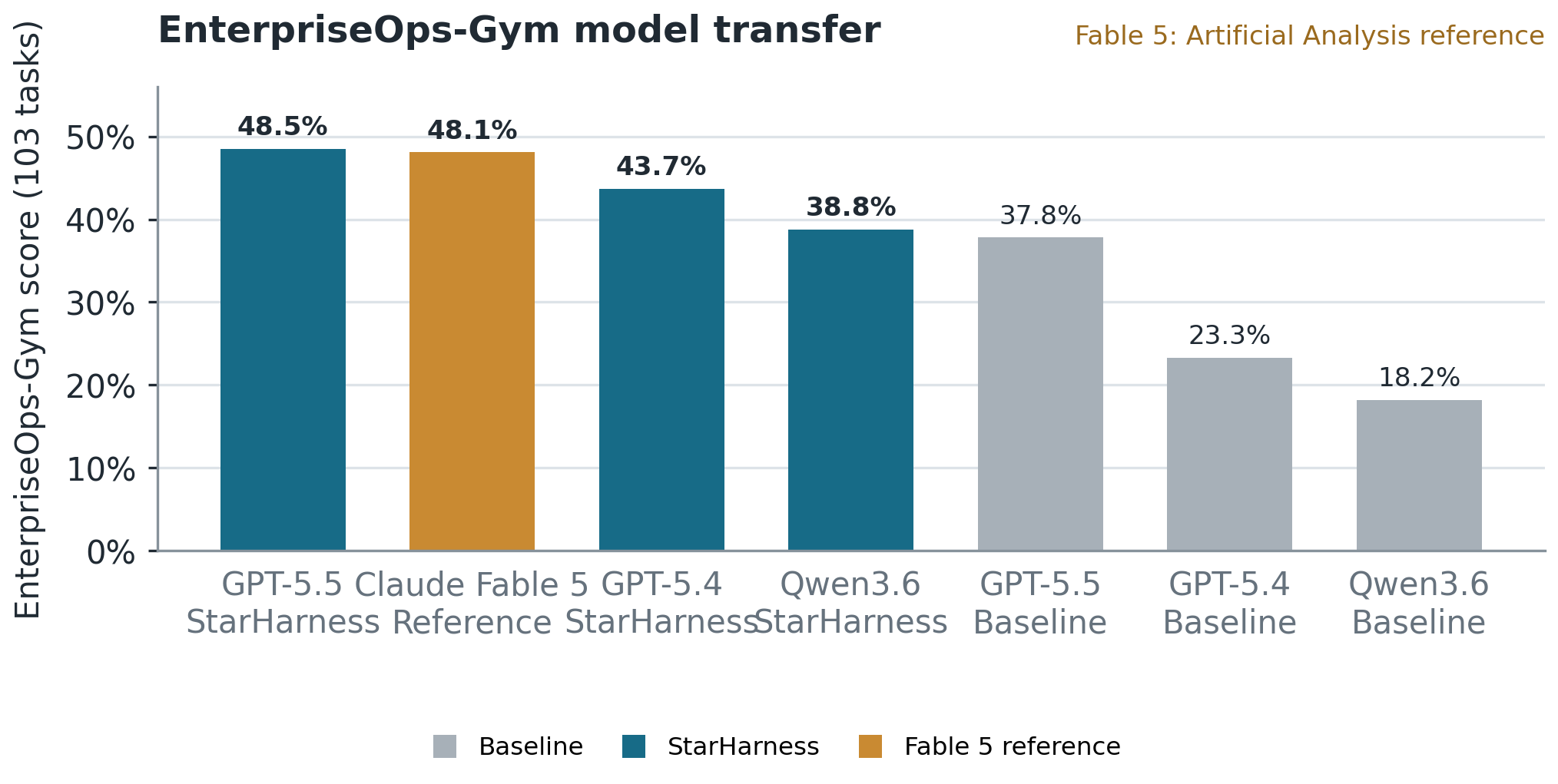}
\caption{EnterpriseOps-Gym model-transfer leaderboard. The Claude Fable 5 score is an external Artificial Analysis reference and is not part of the controlled baseline--StarHarness comparison.}
\label{fig:eops-model-transfer}
\end{figure}

For context, the EnterpriseOps-Gym run also reports external Claude reference scores of 48.1\% (Fable 5), 35.9\% (Sonnet 5), and 35.5\% (Opus 4.8 max), shown in Figure~\ref{fig:eops-model-transfer}. These reference runs do not use the controlled baseline--StarHarness comparison.

\subsection{Held-Out Generalization}

Table~\ref{tab:holdout} summarizes GPT-5.4 generalization across the three benchmarks.

\begin{table}[!htbp]
\caption{GPT-5.4 generalization summary. Values are absolute gains in percentage points.}
\label{tab:holdout}
\centering
\small
\begin{tabular}{@{}llrr@{}}
\toprule
Benchmark & Model & Evolution set & Held-out \\
\midrule
ITBench & GPT-5.4 & +45.0 & +31.7 \\
EnterpriseOps-Gym & GPT-5.4 & +22.0 & +15.1 \\
AutomationBench & GPT-5.4 & +23.0 & +29.3 \\
\bottomrule
\end{tabular}
\end{table}
\FloatBarrier

\section{Analysis: What Harness Evolution Learns}

Across the three evolution runs, 21 patches were accepted: 4 for ITBench, 12 for EnterpriseOps-Gym, and 5 for AutomationBench. On EnterpriseOps-Gym, 8 accepted patches came from the tree-search exploration stage and 4 from the subsequent hill-climbing stage. The tree stage exposed interacting schema, prompt, and tool failures; hill climbing then added targeted linkage and grounding fixes. Across benchmarks, the edits addressed three kinds of model--environment friction.

\textbf{Interface repair.}
EnterpriseOps-Gym evolution repaired MCP argument handling, preserved compound schemas, pruned misleading fields, and added linkage and self-reference cues. These changes made the existing environment interface more usable without changing the task data or verifier. AutomationBench similarly gained structured row operations that replaced fragile raw spreadsheet edits.

Codex provides a close reference on EnterpriseOps-Gym: it scores 41.7\%, compared with 43.7\% for StarHarness. Its default MCP preprocessing normalizes and compacts schemas, reducing invalid calls to the strict Docker-backed server, which rejects explicit \texttt{null}s and empty values when they violate the schema. StarHarness learned complementary repairs: it narrows and enriches schemas, then strips null, empty, and placeholder arguments before calls reach the server.

\textbf{Environment conventions.}
The evolved harness made implicit operational rules explicit. Examples include the EnterpriseOps-Gym execution contract, coupled priority and impact/urgency updates, and the requirement to preserve relationship fields. In AutomationBench, system guidance anchored relative dates to the sandbox clock and added a triage step before mutations. The harness encoded these procedures while the model weights stayed fixed.

\textbf{Operational knowledge and search compression.}
Several patches moved repeatable work out of open-ended reasoning. ITBench gained a forensics overview that ranks candidate upstream causes from observed evidence. AutomationBench gained date and finance calculators for deterministic temporal and numerical operations. These changes encode operational knowledge and compress search; they expose derived information from the task environment without accessing labels or verifier state.

\subsection{Trajectory Analysis}
\label{sec:trajectory}

Each table reports the baseline, evolved harness, and absolute change for the available paired records.

\textbf{ITBench.} Table~\ref{tab:itbench-traces} reports full-benchmark score and execution traces over all 40 tasks. Evolution increased task score from 40.0\% to 75.0\%, reduced turns and false positives, and increased true positives while using a comparable number of shell calls.

\begin{table}[!htbp]
\caption{ITBench (GPT-5.4) full-benchmark score, execution-trace, and estimated API-cost comparison over all 40 tasks.}
\label{tab:itbench-traces}
\centering
\small
\begin{tabular}{@{}lrrr@{}}
\toprule
Metric & Baseline & Evolved & $\Delta$ \\
\midrule
Full-benchmark task score & 40.0\% & 75.0\% & +35.0 pp \\
Turns per task & 25.2 & 22.1 & $-$3.1 \\
Shell calls & 49.8 & 46.7 & $-$3.1 \\
Cost per task & \$3.26 & \$2.70 & $-$17\% \\
False positives & 0.79 & 0.33 & $-$0.46 \\
True positives & 0.45 & 0.78 & +0.33 \\
\bottomrule
\end{tabular}
\end{table}
\FloatBarrier

The evolved agent reached more accurate conclusions with fewer turns while inspecting a comparable amount of raw evidence. Regressions occurred when upstream search continued after a proximate cause should have ended it.

\textbf{EnterpriseOps-Gym.} Table~\ref{tab:eops-traces} reports task success and execution traces over the full benchmark. Evolution shortened workflows and increased verifier completion.

\begin{table}[!htbp]
\caption{EnterpriseOps-Gym (GPT-5.4) full-benchmark task-success, execution-trace, and estimated API-cost comparison over 103 tasks.}
\label{tab:eops-traces}
\centering
\small
\begin{tabular}{@{}lrrr@{}}
\toprule
Metric & Baseline & Evolved & $\Delta$ \\
\midrule
Full-benchmark task success & 23.3\% & 43.7\% & +20.4 pp \\
Turns per task & 18.12 & 9.87 & $-$8.25 \\
Tool calls & 29.53 & 16.83 & $-$12.70 \\
Cost per task & \$1.23 & \$0.58 & $-$53\% \\
Verifier pass rate & 34.5\% & 72.8\% & +38.3 pp \\
\bottomrule
\end{tabular}
\end{table}
\FloatBarrier

Schema and argument repairs target tool selection. Execution and finish contracts target incomplete workflows. Self-reference and linkage cues preserve state across dependent mutations.

\textbf{AutomationBench.} Table~\ref{tab:ab-traces} reports the full GPT-5.4 execution-record comparison. On the full GPT-5.4 task set, evolution reduced unsafe execution and improved partial completion.

\begin{table}[!htbp]
\caption{AutomationBench (GPT-5.4) execution-record and estimated API-cost comparison on 100 tasks.}
\label{tab:ab-traces}
\centering
\small
\begin{tabular}{@{}lrrr@{}}
\toprule
Metric & Baseline & Evolved & $\Delta$ \\
\midrule
Domain objective score & 57.1\% & 83.2\% & +26.1 pp \\
Mean partial credit & 67.3\% & 86.1\% & +18.8 pp \\
Turns per task & 16.35 & 11.98 & $-$4.37 \\
Cost per task & \$0.14 & \$0.10 & $-$29\% \\
Tasks with guardrail violations & 20 & 4 & $-$16 \\
Total guardrail violations & 33 & 4 & $-$29 \\
Zero-score tasks & 24 & 6 & $-$18 \\
\bottomrule
\end{tabular}
\end{table}
\FloatBarrier

The harness places triage before mutation, anchors dates to the sandbox clock, and delegates arithmetic and spreadsheet operations to deterministic tools. We cannot isolate the contribution of any individual tool from these records.

Across all three benchmarks, the accepted edits repaired interfaces, exposed environment conventions, encoded operational knowledge, or compressed search over available evidence. We cannot isolate the causal contribution of individual patches from the paired comparisons.

\section{Conclusion}

Our results show evidence that a fixed model can improve substantially when its surrounding harness is adapted to the environment. Across ITBench, EnterpriseOps-Gym, and AutomationBench, StarHarness learned repairs to tool interfaces, environment conventions, operational knowledge, and search-compression aids. These changes improved full-benchmark scores, generalized to tasks excluded from evolution, and transferred across GPT and Qwen model families without re-evolving the harness. Trace analysis associates the gains with shorter workflows, fewer false-positive diagnoses, and fewer unsafe executions in the settings where those measurements were available.

The results suggest that harness evolution is a practical complement to model scaling for stateful enterprise agents. The search changes how a fixed model retrieves records, calls tools, preserves dependencies, and verifies side effects. Future direction is to co-evolve the harness and model weights through reinforcement learning, allowing the scaffold and policy to specialize jointly to enterprise interaction protocols. This could produce smaller, more efficient enterprise models and test whether joint harness--weight optimization can match or exceed larger models at lower inference cost.

\bibliographystyle{servicenow}
\bibliography{references}

\end{document}